\documentclass{article}
\usepackage{graphicx} 
\usepackage[a4paper, margin=1.4in]{geometry}  
\usepackage{hyperref}

\title{Addressing a fundamental limitation in deep vision models: lack of spatial attention}

\author{Ali Borji \\
aliborji@gmail.com}
\date{June 2024}

\begin{document}

\maketitle

\begin{abstract}
The primary aim of this manuscript is to underscore a significant limitation in current deep learning models, particularly vision models. Unlike human vision, which efficiently selects only the essential visual areas for further processing, leading to high speed and low energy consumption, deep vision models process the entire image. In this work, we examine this issue from a broader perspective and propose two solutions that could pave the way for the next generation of more efficient vision models. In the first solution, convolution and pooling operations are selectively applied to altered regions, with a change map sent to subsequent layers. This map indicates which computations need to be repeated. In the second solution, only the modified regions are processed by a semantic segmentation model, and the resulting segments are inserted into the corresponding areas of the previous output map. The code is available at \url{https://github.com/aliborji/spatial_attention}.




\end{abstract}

\section{Motivation}

\begin{figure}
    \centering
    \includegraphics[width=0.45\linewidth]{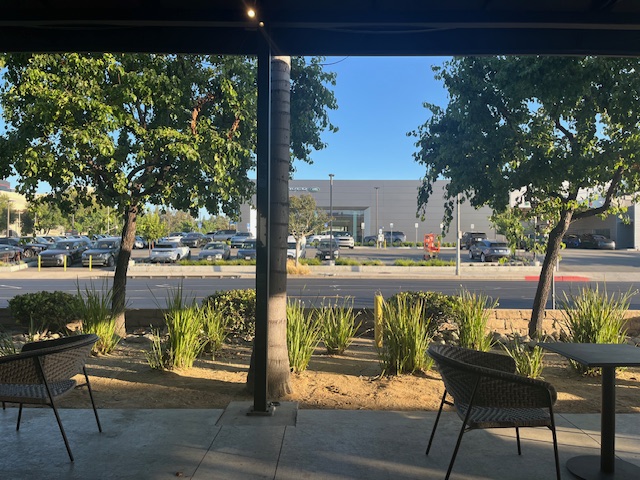}
    \includegraphics[width=0.45\linewidth]{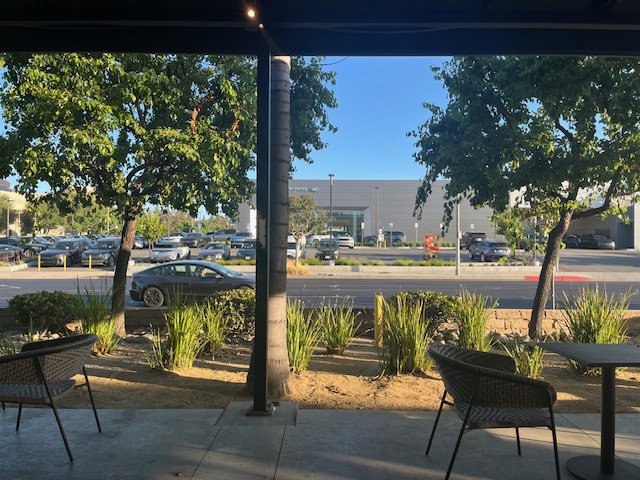}
    \caption{In real-world visual content, not much changes; the majority of the scene is often static. For example, in this image, a car has entered the scene and perhaps some tree branches have moved slightly due to the wind, but the rest of the scene remains largely unchanged. left: t - 1, right: t.}
    \label{fig:teaser}
\end{figure}

The visual world around us is dynamic, and we rarely see the exact same image twice due to variations in lighting and other factors. Similarly, neural activity is not identical even when exposed to the same input. However, not everything in the visual world changes, and often only a small portion of the input varies over short periods (Figure~\ref{fig:teaser}). Our visual system has evolved to efficiently address this by selectively focusing on and processing important regions of interest. In contrast, deep vision models lack this capability. While there have been some ad-hoc approaches to address this issue, they are not inherent to the models. The main problem lies in operations such as convolution ({\bf nn.Conv2d}), which are applied to the entire image without the ability to selectively skip parts of it at the hardware level. We argue that this is a major limitation and propose potential solutions for researchers to explore in the future to address this problem.

Convolutional neural networks and vision transformers lack this selective processing capability. Although various attention mechanisms have been proposed, they do not perform spatial attention. In transformers, attention operates more like feature-based attention, as described in the attention literature, rather than spatial attention. 

In the first proposed approach, computation is performed on demand. One advantage of this method is that it can be applied solely during inference. The model can be trained using a GPU and then optimized using this approach to enhance inference efficiency. In the second approach, only the modified regions are processed by a semantic segmentation model, and the resulting segments are inserted into the corresponding areas of the previous output map.

\section{Related work} 
Visual attention is the cognitive process of selectively focusing on one aspect of the environment while ignoring others~\cite{itti2001computational,borji2012state}. This is essential because the human brain cannot process all visual information simultaneously. There are two main types of attention: 1) Goal-Driven Top-Down Attention: This intentional type is controlled by an individual's goals and expectations. For example, searching for a friend in a crowded place, 2) Bottom-up Attention: This automatic type is triggered by sudden or prominent stimuli, such as a loud noise or a bright flash. The primary purpose of attention is to conserve computational resources by enabling an agent to focus on the most important, task-relevant items and relay them to higher visual areas that require more computational effort. In our visual system, several mechanisms support this function. These range from the hardware aspect of moving the fovea around to specialized mechanisms and circuitry that generate a saliency map to prioritize scene elements for further processing.

The closest concept to this work is event cameras~\cite{amir2017low}, which aim to process only the content that has changed at the hardware level by using specially designed cameras (See also~\cite{klotz2025minimalist}). In video surveillance, some systems efficiently detect and focus on frames with notable movement or changes, enhancing surveillance effectiveness. However, while these approaches skip frames, they still process the entire image when they do choose to process a frame. In contrast, the proposed approach processes only specific regions of the image, providing finer granularity compared to previous methods, although it requires some memory to save previous outputs. 

Handling sparse data in deep learning, especially in convolutional neural networks (CNNs), often involves specialized architectures designed specifically for sparse inputs. One such approach is sparse convolutions, where the operations are limited only to non-zero elements, thereby skipping the unnecessary computation associated with dense convolutions. Submanifold Sparse Convolutions (SSC)~\cite{graham2017submanifold}, for instance, are particularly effective in maintaining sparsity by ensuring that only active (non-zero) features in each layer remain sparse throughout the network. This approach has proven useful in tasks like 3D object detection, where input data is inherently sparse (e.g., LiDAR point clouds). Another method is coordinate-based convolutions, which directly work with sparse coordinate inputs and avoid dense data representation altogether. Additionally, models like Sparse R-CNN~\cite{sun2021sparse} focus on sparse proposals to efficiently handle object detection by leveraging sparse inputs and predictions without requiring exhaustive feature maps. These methods are key in fields where data is naturally sparse, such as 3D vision and graph-based tasks, allowing CNNs to process information more efficiently by focusing only on relevant data points while maintaining performance.

Saliency methods, which highlight important regions behind model decisions, are primarily used for explainability purposes (e.g.,~\cite{zhou2016learning}). These methods are different from saliency models that attempt to select a subset of image or video or predict eye movements~\cite{borji2012state}. So far, models of saliency (the latter type) and visual recognition have not been integrated to create a model that natively supports both.

Additionally, there are approaches that attempt to compress models or prune weights to make them faster~\cite{cheng2017survey}. However, these methods are not directly related to this work, as our focus is on pruning irrelevant or less important content rather than weights. Other potentially related areas include spiking neural networks~\cite{tavanaei2019deep} and predictive coding~\cite{spratling2017review,huang2011predictive}.

\begin{figure}[t]
    \centering
    \includegraphics[width=0.8\linewidth]{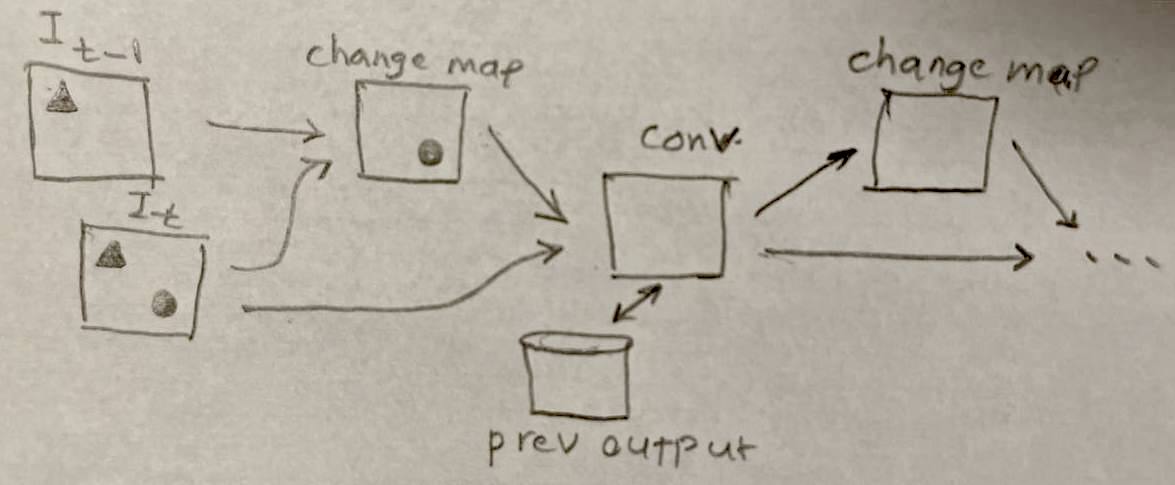}
    \caption{Illustration of the Basic Idea: First, a change map is computed from subsequent frames ($I_{t-1}$ and $I_{t}$). This change map is sent to the first convolution layer, which updates the values in its previous output only for the changed regions. Knowing which regions have been updated, this layer generates its own change map and sends it to the next layer, and so on. Each layer maintains its own memory to avoid redundant calculations. }
    \label{fig:idea}
\end{figure}

\section{Potential solution I}

In this approach, convolution and pooling operations are selectively applied to altered regions, with a change map sent to subsequent layers. This map informs those layers about which computations need to be repeated. Each layer communicates changes so the next layer knows what it needs to recompute, and this process continues until the final layer. To achieve this, each layer must remember its last computation to avoid redundant processing.

The basic idea is illustrated in Figure~\ref{fig:idea}. First, a change map is computed from subsequent frames (e.g., $|I_t - I_{t-1}|$). This change map is sent to the first convolution layer, which updates its previous output\footnote{For the very first image, the previous output of the first convolution layer is set to zeros. Please see the code.} values only for the changed regions\footnote{Note that there is no need to compare the current output with the previous output to calculate the change map, although this is an option as well.}. Knowing which regions are updated, it can compute a change map for itself and send it to the next layer, and so on. Each layer has its own memory, which means extra memory (in addition to model weights) is needed for housekeeping.

The initial change map is set to all ones\footnote{A similar initialization can be done by adding a blank frame at the beginning of the sequence.}. At the frame level, frames are subtracted from each other ($np.abs(I_{t} - I_{t-1})$). Convolution is implemented by looping over spatial locations. If there is enough change (determined by L1 or L2 norm greater than threshold $\tau$) inside a receptive field (RF), that RF is processed; otherwise, it is discarded\footnote{In practice, a layer uses its previously computed output at time $t-1$ and only updates some elements within it.}. This results in significant computational savings, as the filters are not applied to unchanged locations. The conv layer keeps track of changes in its output map and generates a binary map where a 1 indicates a change. This map is sent to the subsequent pooling and convolution layers, and each layer saves its output for future use.

\begin{figure}
    \centering
    \includegraphics[width=\linewidth]{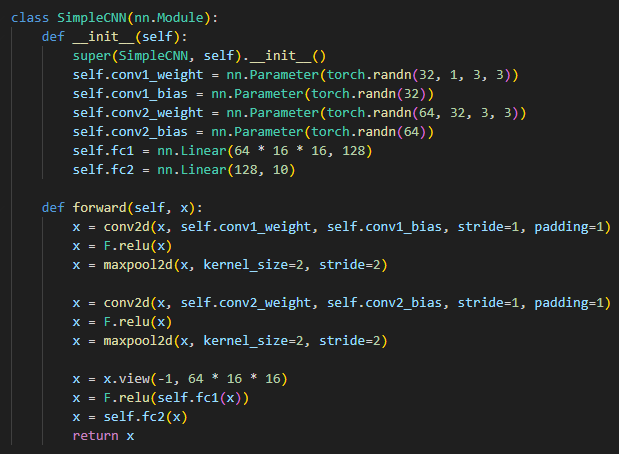}
    \includegraphics[width=\linewidth]{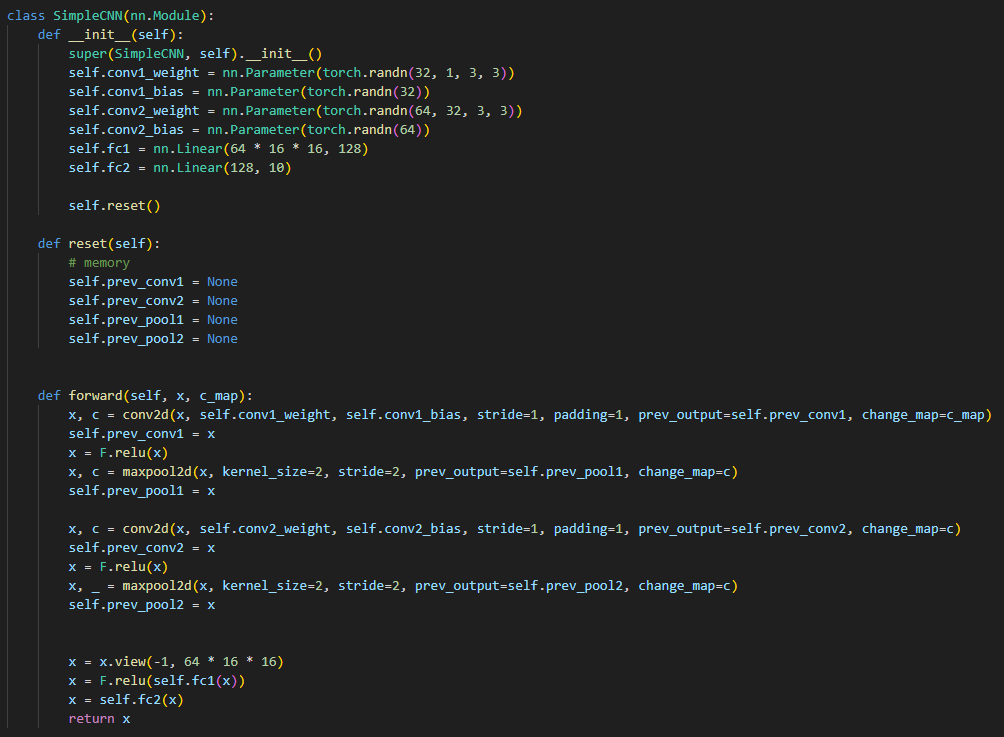}
    \caption{Top: Architecture of CNN1 and CNN2, Bottom: CNN3 Architecture.}
    \label{fig:cnns}
\end{figure}

Three types of CNNs (Figure~\ref{fig:cnns}) are compared. CNN1 is the classic version that uses `nn.Conv2d`. CNN2 sequentially applies convolution to image regions. CNN3 is similar to CNN2 but skips regions that have not changed. Notice that our implementation here (CNN3) is even slower than using nn.conv2d on CPU (CNN1). The main point here is that sequential implementation on CPUs can be modified to save energy and to increase speed. Therefore, further investigation is needed to determine how this concept can be adopted for parallel processing on multi-core CPUs and GPUs.


moving camera!!

\subsection{Experiments and results}

The 28 x 28 MNIST digits, both during training and inference, were placed at the center of a black 64 x 64 image. We trained a simple CNN, referred to as CNN1 as illustrated in Figure~\ref{fig:cnns}, on a GPU with a batch size of 32. Since our primary focus is on the inference stage, we then loaded the weights into a model residing on a CPU\footnote{We used a 3100 MHz Intel(R) Xeon(R) CPU with 8 cores and 64 GB RAM}. A single frame was processed at a time (i.e., batch size = 1). We conducted two experiments as detailed below. The results are presented in Table~\ref{table:results}.

\textbf{Experiment I: Processing repeated versions of the same image}

\begin{figure}
    \centering

    \includegraphics[width=0.07\linewidth]{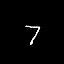}
    \includegraphics[width=0.07\linewidth]{figs/img_0.jpg}
    \includegraphics[width=0.07\linewidth]{figs/img_0.jpg}
    \includegraphics[width=0.07\linewidth]{figs/img_0.jpg}
    \includegraphics[width=0.07\linewidth]{figs/img_0.jpg}
    \includegraphics[width=0.07\linewidth]{figs/img_0.jpg}
    \includegraphics[width=0.07\linewidth]{figs/img_0.jpg}
    \includegraphics[width=0.07\linewidth]{figs/img_0.jpg}
    \includegraphics[width=0.07\linewidth]{figs/img_0.jpg}
    \includegraphics[width=0.07\linewidth]{figs/img_0.jpg}

    \includegraphics[width=0.07\linewidth]{figs/img_0.jpg}
    \includegraphics[width=0.07\linewidth]{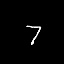}
    \includegraphics[width=0.07\linewidth]{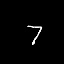}
    \includegraphics[width=0.07\linewidth]{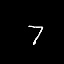}
    \includegraphics[width=0.07\linewidth]{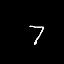}
    \includegraphics[width=0.07\linewidth]{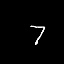}
    \includegraphics[width=0.07\linewidth]{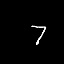}
    \includegraphics[width=0.07\linewidth]{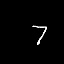}
    \includegraphics[width=0.07\linewidth]{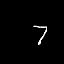}
    \includegraphics[width=0.07\linewidth]{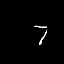}
    
    \caption{Sequences of Images Presented to the CNNs: Top row (Exp I): The digits remain unchanged. Bottom row (Exp II): A digit is progressively shifted 1 pixel to the right.}
    \label{fig:mnist_samples}
\end{figure}

In this experiment, we ran three models on 11 images. Each image is repeated 10 times (110 images in total). The CNN1 model proved to be the fastest because it uses `nn.Conv2d`, which is a parallel implementation on CPU cores. The aim here is to demonstrate that significant computation can be saved when there is no change in the image. Most of the processing is done on the first frame, which is then reused for subsequent frames. This is why the processing time for CNN3 is nearly 1/10th of CNN2. The inputs and results are illustrated in the first rows of Figure~\ref{fig:mnist_samples} and Table~\ref{table:results}, respectively.

\textbf{Experiment II: Processing shifted versions of the same image}
This experiment is similar to Experiment I, except each of the 11 digits is shifted rightward by one pixel at a time, resulting in 110 images in total. This method causes some regions of the image to remain the same while others change. As a result, CNN3 is slower here compared to its speed in Experiment I because it needs to recompute more information due to the increased amount of changed content.

In terms of accuracy, CNN1 and CNN2 are equivalent since their implementations are essentially the same. CNN3, however, can exhibit different performance based on the amount of change ($\tau$). Smaller values of $\tau$ lead to more computation and higher accuracy, and vice versa. Overall, the CNNs performed similarly to each other, although they were less accurate compared to Experiment I, due to pixel shifts. The inputs and results are illustrated in the second rows of Figure~\ref{fig:mnist_samples} and Table~\ref{table:results}, respectively.

\begin{figure}[htbp]
    \centering
    
    \includegraphics[width=0.07\linewidth]{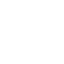}
    \includegraphics[width=0.07\linewidth]{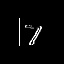}
    \includegraphics[width=0.07\linewidth]{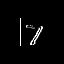}
    \includegraphics[width=0.07\linewidth]{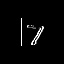}
    \includegraphics[width=0.07\linewidth]{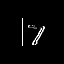}
    \includegraphics[width=0.07\linewidth]{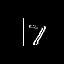}
    \includegraphics[width=0.07\linewidth]{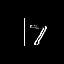}
    \includegraphics[width=0.07\linewidth]{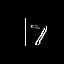}
    \includegraphics[width=0.07\linewidth]{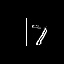}\\
    \includegraphics[width=0.07\linewidth]{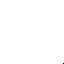}
    \includegraphics[width=0.07\linewidth]{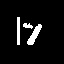}
    \includegraphics[width=0.07\linewidth]{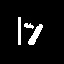}
    \includegraphics[width=0.07\linewidth]{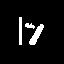}
    \includegraphics[width=0.07\linewidth]{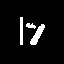}
    \includegraphics[width=0.07\linewidth]{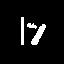}
    \includegraphics[width=0.07\linewidth]{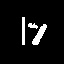}
    \includegraphics[width=0.07\linewidth]{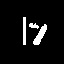}
    \includegraphics[width=0.07\linewidth]{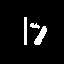}\\
    \includegraphics[width=0.07\linewidth]{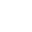}
    \includegraphics[width=0.07\linewidth]{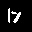}
    \includegraphics[width=0.07\linewidth]{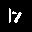}
    \includegraphics[width=0.07\linewidth]{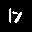}
    \includegraphics[width=0.07\linewidth]{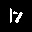}
    \includegraphics[width=0.07\linewidth]{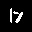}
    \includegraphics[width=0.07\linewidth]{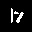}
    \includegraphics[width=0.07\linewidth]{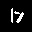}
    \includegraphics[width=0.07\linewidth]{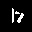}\\
    \includegraphics[width=0.07\linewidth]{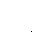}
    \includegraphics[width=0.07\linewidth]{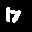}
    \includegraphics[width=0.07\linewidth]{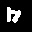}
    \includegraphics[width=0.07\linewidth]{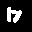}
    \includegraphics[width=0.07\linewidth]{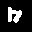}
    \includegraphics[width=0.07\linewidth]{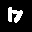}
    \includegraphics[width=0.07\linewidth]{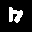}
    \includegraphics[width=0.07\linewidth]{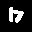}
    \includegraphics[width=0.07\linewidth]{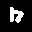} \\
    \includegraphics[width=0.07\linewidth]{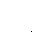}
    \includegraphics[width=0.07\linewidth]{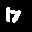}
    \includegraphics[width=0.07\linewidth]{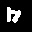}
    \includegraphics[width=0.07\linewidth]{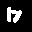}
    \includegraphics[width=0.07\linewidth]{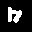}
    \includegraphics[width=0.07\linewidth]{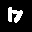}
    \includegraphics[width=0.07\linewidth]{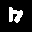}
    \includegraphics[width=0.07\linewidth]{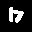}
    \includegraphics[width=0.07\linewidth]{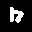}

    \caption{Top row: The change map (64 x 64) showing the effect of shifting the image 1 pixel to the right. The subsequent rows, in order, show the change maps at each layer: 1st conv layer (64 x 64), 1st pooling layer (32 x 32), 2nd conv layer (32 x 32), and 2nd pooling layer (32 x 32). Each change map represents the output of a layer and is passed to the next layer to indicate what needs to be reused and what requires recomputation.}
    \label{fig:shift_diff1_maps}
\end{figure}

\begin{figure}
    \centering
    \includegraphics[width=0.6\linewidth]{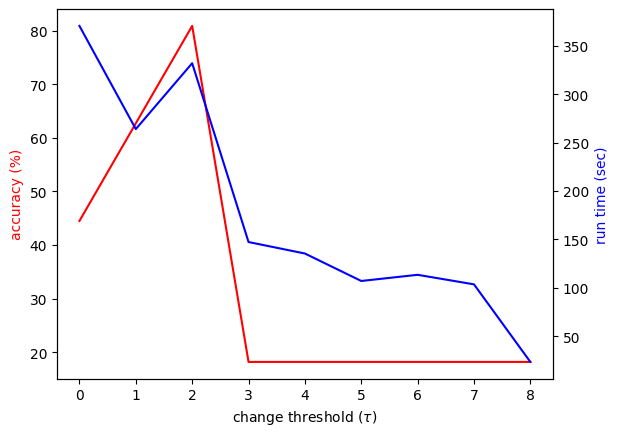}
    \caption{Accuracy and run time as a function of the difference threshold $\tau$. As $\tau$ increases, implying a more stringent condition for considering a pixel or patch as changed, the run time decreases.}
    \label{fig:diff_analysis}
\end{figure}

The change maps for the input images and the network layers are displayed in Figure~\ref{fig:shift_diff1_maps}. As we increased the change threshold $\tau$, the accuracy decreased while the speed increased. This relationship is illustrated in Figure~\ref{fig:diff_analysis}.

\begin{table}[t]
\centering
\begin{tabular}{|c|c|c|c|}
    \hline
     & \textbf{CNN1} & \textbf{CNN2} & \textbf{CNN3} \\
    \hline
    \textbf{Exp I: repeated image (110)} & 90.91, 0.212 & 90.91, 2174.75 & 90.91, 213.2 \\
    \hline
    \textbf{Exp II: shifted image (110)} & 43.63, 0.172 & 43.63, 1664.5 & 44.5, 370.5 \\
    \hline
\end{tabular}
\caption{Comparison of accuracy and run time for the three CNNs. Run time (the second number in each cell) is measured in seconds.}
\label{table:results}
\end{table}

\section{Potential solution II}

\begin{figure}
    \centering
    \includegraphics[width=1\linewidth]{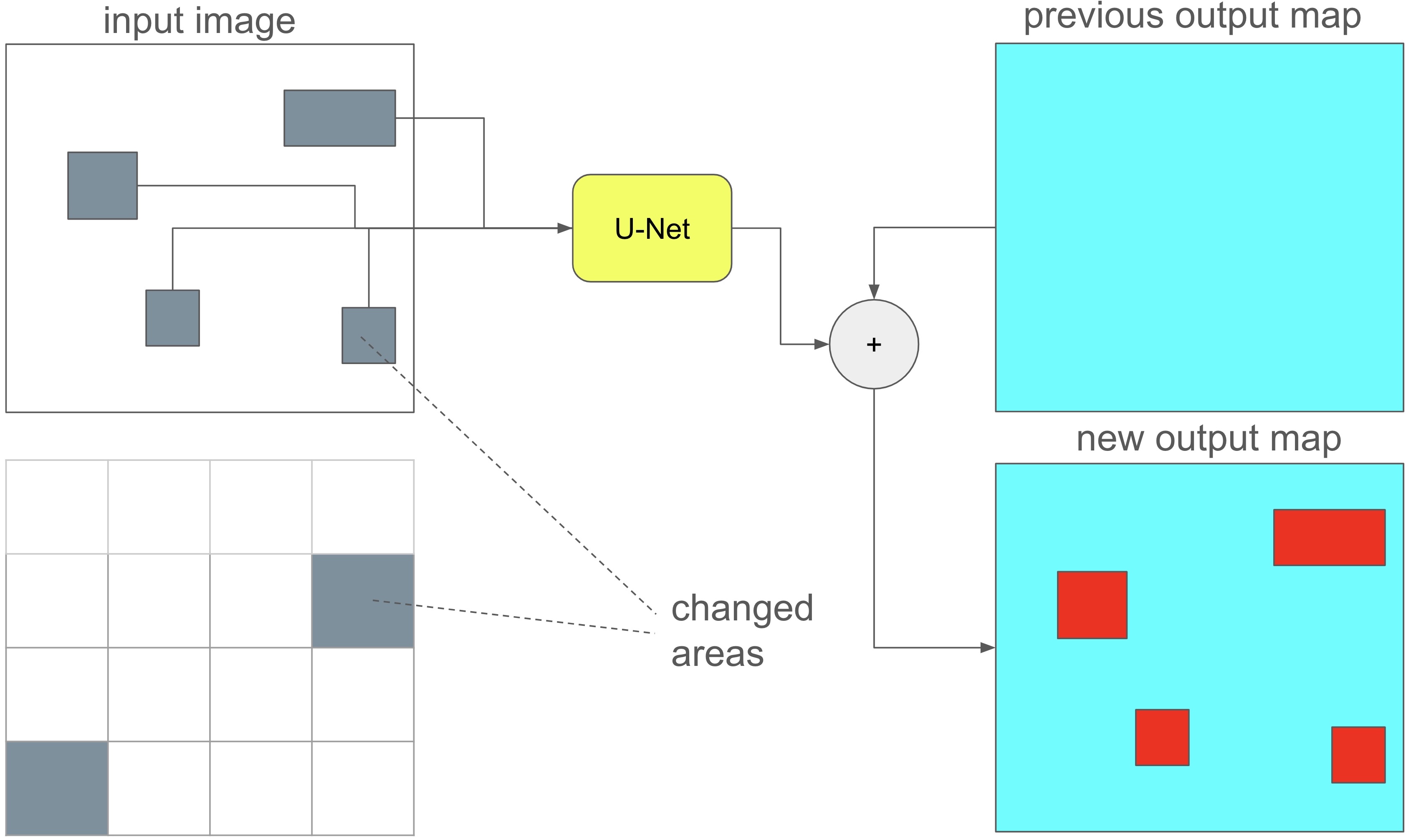}
    \caption{The schematic diagram of the second proposed solution highlights that only the modified regions (bounded by boxes fitting the changed areas, such as the top-left region or altered grid locations) are processed by U-Net. The segmentation results for these regions are then merged into the previous output map.}
    \label{fig:unet}
\end{figure}

The second method involves semantic segmentation, often regarded as a supertask. This is because the capability to label each individual pixel can generalize to other tasks, such as object recognition and object detection\footnote{Instance segmentation, however, is even more comprehensive since it distinguishes between instances of the same object.}. The idea here is to feed only the modified regions into the segmentation model. A benefit of some models, especially those based on U-Net architectures~\cite{ronneberger2015u}, is that they can process inputs of any resolution. After obtaining the output, it can be accurately positioned in the prior output map. Please see Fig.~\ref{fig:unet}.

\begin{figure}
    \centering
    \includegraphics[width=1\linewidth]{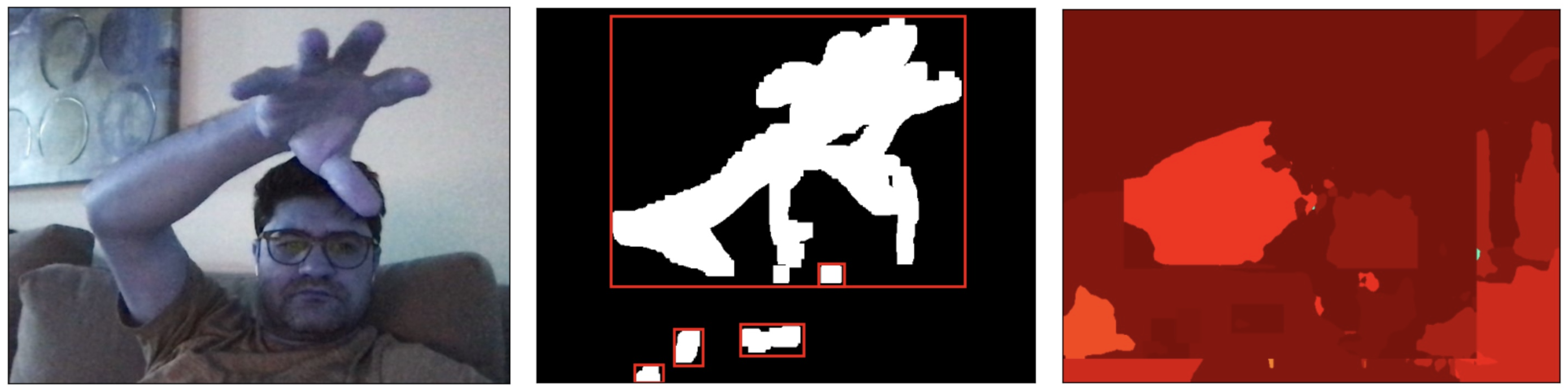}
    \caption{The approach for the second proposed solution is illustrated as follows: first, a binary change map is generated to highlight only the altered pixels, a computationally inexpensive process. Next, bounding boxes are fitted to these changed regions, and the model is applied exclusively to those areas. The segmentation map is shown here for illustrative purposes only, as the model used has not been specifically trained for human faces and bodies, which explains the lack of accuracy. The change map is computed similar to the first approach shown in Figure~\ref{fig:idea}.}
    \label{fig:semantic}
\end{figure}

Figure~\ref{fig:semantic} provides a more detailed illustration. In the middle panel, the regions with changes from the previous frame (e.g., hand movement) are highlighted. Through image processing operations like dilation and region labeling, bounding boxes are fitted to these changed areas. The model is then applied solely to the image regions within these bounding boxes, and the results are then overlaid onto the previous segmentation map. Since typically only a small portion of the image changes, this approach saves significant computation. Although this method may result in a minor accuracy drop due to the lack of global context when processing local regions, it is not a primary concern here. We focus on presenting the method, with the possibility that future segmentation models could incorporate global context even in localized regions. Our preliminary, unoptimized results show an average of 2.44 FPS using this approach, compared to 1.69 FPS when the full frame is processed by the model. For more details, please refer to the accompanying code.

\section{Conclusion}

We highlighted a key issue with existing deep learning approaches and proposed solutions that can be integrated into current models or used to design new models with inherent attention and memory mechanisms. This work serves as a proof of concept and can be applied to other problems such as object detection, scene segmentation, and action recognition. It also has the potential to help address adversarial examples~\cite{borji2021contemplating}. This method is particularly effective when the input has high resolution.

Future work could explore techniques for handling sparse data, such as those discussed in the related works section. While this study focused on CPU optimization, extending these methods for GPU implementation is a promising direction for further research.

Our visual system is far more sophisticated and energy-efficient than the most advanced deep learning models available today. Specifically, our early visual system performs extensive preprocessing tasks such as saliency computation, foveation, gaze control, shape and texture processing~\cite{geirhos2018imagenet}, and background subtraction. These mechanisms not only enhance processing speed and energy efficiency but also provide robustness against input distortions and improve generalizability across various tasks. We should certainly draw inspiration from our own visual system to develop better deep learning models.

\section{Appendix}

\begin{figure}[t]
    \centering
    \includegraphics[width=0.8\linewidth]{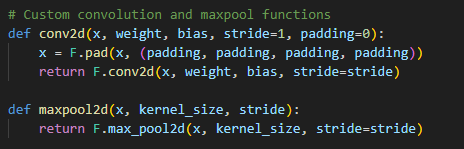}

    \caption{CNN modules for the CNN1 model}
    \label{fig:cnn1-modules}
\end{figure}

\begin{figure}[htbp]
    \centering
    \includegraphics[width=1\linewidth]{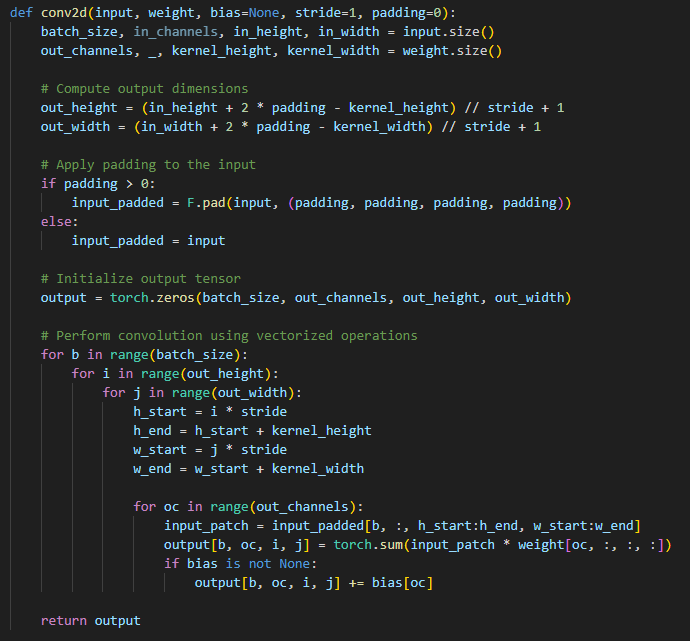}
     \includegraphics[width=1\linewidth]{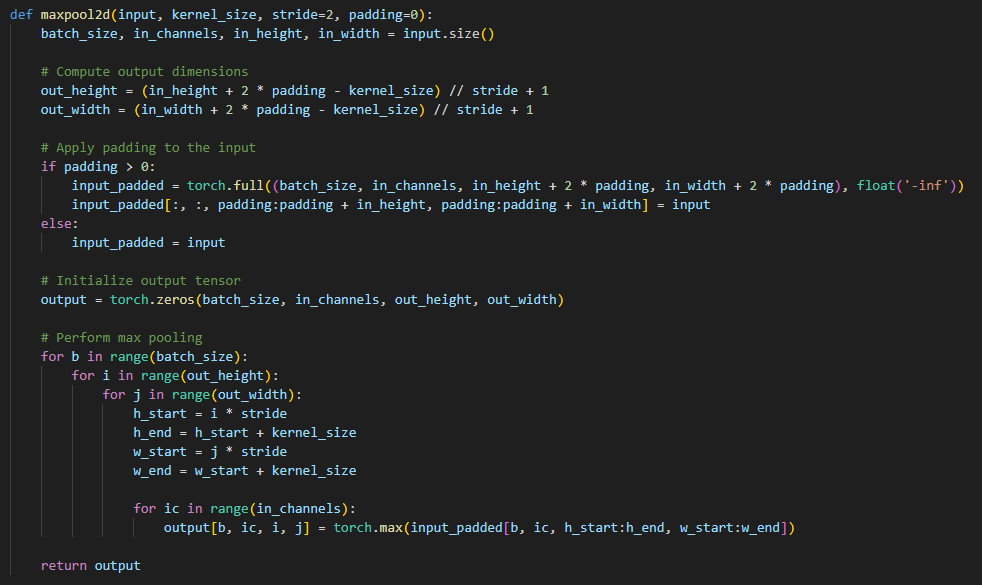}
    \caption{CNN modules for the CNN2 model}
    \label{fig:cnn2-modules}

\end{figure}

\begin{figure}
    \centering
    \includegraphics[width=1\linewidth]{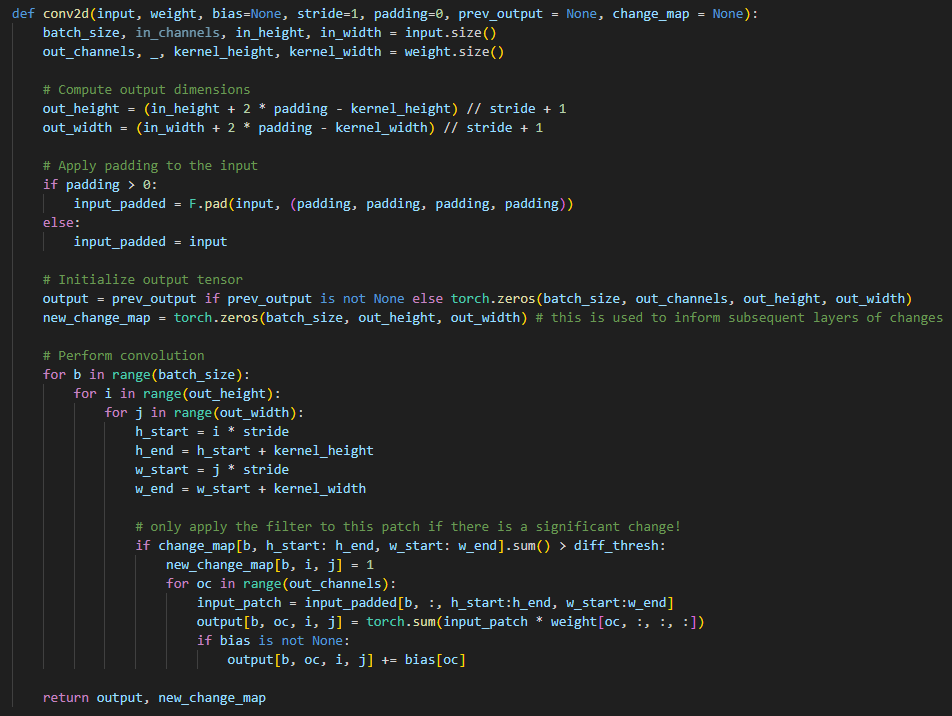}
     \includegraphics[width=1\linewidth]{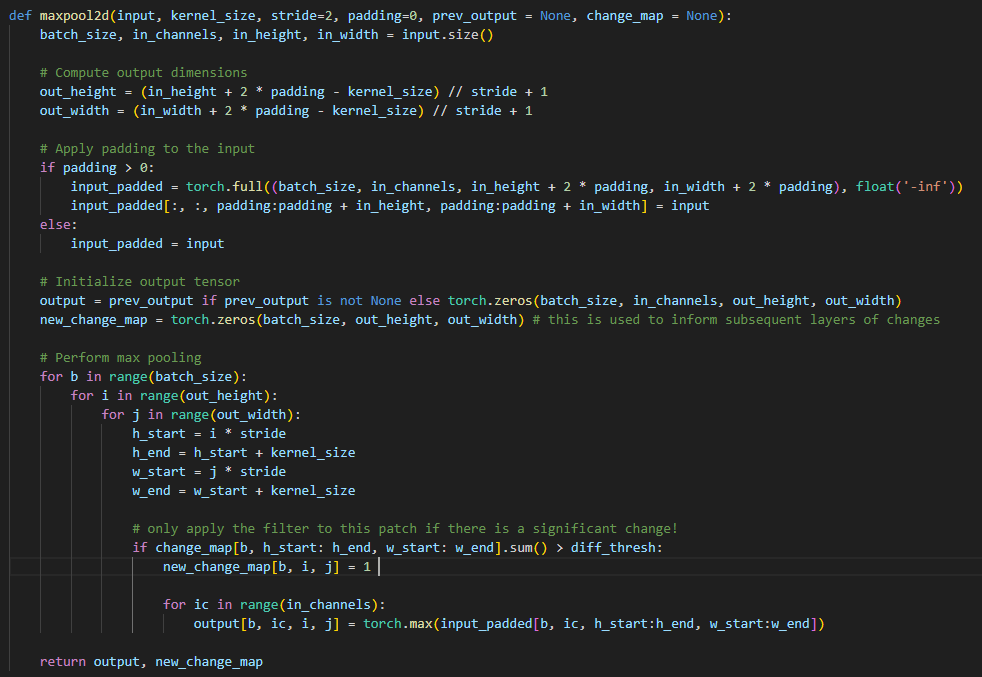}
    \caption{CNN modules for the CNN3 model}
    \label{fig:cnn3-modules}

\end{figure}

Convolution and pooling modules of CNNs are shown in Figures~\ref{fig:cnn1-modules},~\ref{fig:cnn2-modules}, and ~\ref{fig:cnn3-modules}.

\bibliographystyle{plain}
\bibliography{refs}

\begin{thebibliography}{10}

\bibitem{amir2017low}
Arnon Amir, Brian Taba, David Berg, Timothy Melano, Jeffrey McKinstry, Carmelo Di~Nolfo, Tapan Nayak, Alexander Andreopoulos, Guillaume Garreau, Marcela Mendoza, et~al.
\newblock A low power, fully event-based gesture recognition system.
\newblock In {\em Proceedings of the IEEE conference on computer vision and pattern recognition}, pages 7243--7252, 2017.

\bibitem{borji2021contemplating}
Ali Borji.
\newblock Contemplating real-world object classification.
\newblock {\em arXiv preprint arXiv:2103.05137}, 2021.

\bibitem{borji2012state}
Ali Borji and Laurent Itti.
\newblock State-of-the-art in visual attention modeling.
\newblock {\em IEEE transactions on pattern analysis and machine intelligence}, 35(1):185--207, 2012.

\bibitem{cheng2017survey}
Yu~Cheng, Duo Wang, Pan Zhou, and Tao Zhang.
\newblock A survey of model compression and acceleration for deep neural networks.
\newblock {\em arXiv preprint arXiv:1710.09282}, 2017.

\bibitem{geirhos2018imagenet}
Robert Geirhos, Patricia Rubisch, Claudio Michaelis, Matthias Bethge, Felix~A Wichmann, and Wieland Brendel.
\newblock Imagenet-trained cnns are biased towards texture; increasing shape bias improves accuracy and robustness.
\newblock {\em arXiv preprint arXiv:1811.12231}, 2018.

\bibitem{graham2017submanifold}
Benjamin Graham and Laurens Van~der Maaten.
\newblock Submanifold sparse convolutional networks.
\newblock {\em arXiv preprint arXiv:1706.01307}, 2017.

\bibitem{huang2011predictive}
Yanping Huang and Rajesh~PN Rao.
\newblock Predictive coding.
\newblock {\em Wiley Interdisciplinary Reviews: Cognitive Science}, 2(5):580--593, 2011.

\bibitem{itti2001computational}
Laurent Itti and Christof Koch.
\newblock Computational modelling of visual attention.
\newblock {\em Nature reviews neuroscience}, 2(3):194--203, 2001.

\bibitem{klotz2025minimalist}
Jeremy Klotz and Shree~K Nayar.
\newblock Minimalist vision with freeform pixels.
\newblock In {\em European Conference on Computer Vision}, pages 329--346. Springer, 2025.

\bibitem{ronneberger2015u}
Olaf Ronneberger, Philipp Fischer, and Thomas Brox.
\newblock U-net: Convolutional networks for biomedical image segmentation.
\newblock In {\em Medical image computing and computer-assisted intervention--MICCAI 2015: 18th international conference, Munich, Germany, October 5-9, 2015, proceedings, part III 18}, pages 234--241. Springer, 2015.

\bibitem{spratling2017review}
Michael~W Spratling.
\newblock A review of predictive coding algorithms.
\newblock {\em Brain and cognition}, 112:92--97, 2017.

\bibitem{sun2021sparse}
Peize Sun, Rufeng Zhang, Yi~Jiang, Tao Kong, Chenfeng Xu, Wei Zhan, Masayoshi Tomizuka, Lei Li, Zehuan Yuan, Changhu Wang, et~al.
\newblock Sparse r-cnn: End-to-end object detection with learnable proposals.
\newblock In {\em Proceedings of the IEEE/CVF conference on computer vision and pattern recognition}, pages 14454--14463, 2021.

\bibitem{tavanaei2019deep}
Amirhossein Tavanaei, Masoud Ghodrati, Saeed~Reza Kheradpisheh, Timoth{\'e}e Masquelier, and Anthony Maida.
\newblock Deep learning in spiking neural networks.
\newblock {\em Neural networks}, 111:47--63, 2019.

\bibitem{zhou2016learning}
Bolei Zhou, Aditya Khosla, Agata Lapedriza, Aude Oliva, and Antonio Torralba.
\newblock Learning deep features for discriminative localization.
\newblock In {\em Proceedings of the IEEE conference on computer vision and pattern recognition}, pages 2921--2929, 2016.

\end{thebibliography}

\end{document}